\pdfoutput=1

\documentclass[11pt]{article}

\usepackage[]{ACL2023}

\usepackage{graphicx}
\usepackage{times}
\usepackage{latexsym}
\usepackage{array}
\usepackage{ragged2e}

\usepackage[T1]{fontenc}

\usepackage[utf8]{inputenc}

\usepackage{microtype}

\usepackage{inconsolata}

\title{Dissecting vocabulary biases datasets through statistical testing and automated data augmentation for artifact mitigation in Natural Language Inference}

\author{Dat Thanh Nguyen \\
  Department of Computer Science, University of Texas at Austin \\
  \texttt{ndat@utexas.edu} 
  }

\begin{document}
\maketitle
\begin{abstract}

In recent years, the availability of large-scale annotated datasets, such as the Stanford Natural Language Inference and the Multi-Genre Natural Language Inference, coupled with the advent of pre-trained language models, has significantly contributed to the development of the natural language inference domain. However, these crowdsourced annotated datasets often contain biases or dataset artifacts, leading to overestimated model performance and poor generalization. In this work, we focus on investigating dataset artifacts and developing strategies to address these issues. Through the utilization of a novel statistical testing procedure, we discover a significant association between vocabulary distribution and text entailment classes, emphasizing vocabulary as a notable source of biases. To mitigate these issues, we propose several automatic data augmentation strategies spanning character to word levels. By fine-tuning the ELECTRA pre-trained language model, we compare the performance of boosted models with augmented data against their baseline counterparts. The experiments demonstrate that the proposed approaches effectively enhance model accuracy and reduce biases by up to 0.66\% and 1.14\%, respectively.

\end{abstract}

\section{Introduction}

In the last few years, natural language inference (NLI), a sub-domain of natural language processing (NLP), has been an active research field. In NLI, systems are asked to determine whether a hypothesis sentence (h) can be logically inferred (entailed) from a given premise sentence (p) \cite{maccartney2008modeling}. For example, a hypothesis, \textit{"A family of three is at the beach"} would likely to be inferred from the premise of \textit{"A man, woman, and child enjoying themselves on a beach"}. This task has been proved challenging due to the complex nature of natural language, which involves logical reasoning, lexical semantic knowledge, and variations in linguistic expression. Recent years have witnessed significant research efforts in the field of NLI, notably marked by the creation of benchmark datasets such as the Stanford Natural Language Inference (SNLI) corpus \cite{bowman2015large} and the Multi-Genre Natural Language Inference (MNLI) corpus \cite{williams2017broad}. These datasets provide substantial amounts of annotated data, fostering the development and evaluation of NLP models. Additionally, the advent of pre-trained language models has played a key role in enhancing the efficacy of the NLI problem.

Regardless of the usefulness of these large datasets in the training and evaluation of NLI systems, recent works showed that they contain biases or dataset artifacts, i.e., there are spurious patterns present in the datasets that NLI models can exploit to achieve high performance without learning relationships of the premise and hypothesis sentences \cite{poliak2018hypothesis, gururangan2018annotation, tsuchiya2018performance}. The cause of this phenomenon possibly arises from the fact that crowdsourcing annotated datasets often employ a few strategies to obtain different entailment conditions. For example, negation words such as \textit{"not, non, nobody"} are likely associated with contradiction \cite{gururangan2018annotation}. Furthermore, modifying or omitting gender information, and neutralizing gender information by simply counting the number of persons in the premise for neutral pairs are also common practices during the generation of entailment hypotheses. For instance, Table~\ref{tab1} includes three different examples from the SNLI dataset that simply modify the subject word \textit{"people"} to a specific gender word such as \textit{"man"}, or \textit{"woman"} to obtain the neutral hypothesis examples. As a result, NLI models trained on datasets with such artifacts may tend to overestimate their performance and generalize poorly in other settings.

In this work, we focus on investigating dataset artifacts in NLI domain and developing strategies to mitigate these issues. Specifically, we identify dataset artifacts in the SNLI dataset by a novel statistical testing procedure. We show that vocabulary distribution in the hypothesis is the source of biases. We then propose several automatic data augmentation strategies for mitigating artifacts in this dataset. We finally employ the hypothesis-only training approach \cite{gururangan2018annotation} to fine-tune the pre-trained language model of Efficiently Learning an Encoder that Classifies Token Replacements Accurately (ELECTRA) \cite{clark2020electra} to evaluate the efficiency of the proposed data augmentations.

\begin{table*}
\centering
\begin{tabular}{p{6cm}p{6cm}p{2cm}} 
\hline
\textbf{Premise}                                                           & \textbf{Hypothesis}                                          & \textbf{Label}    \\ 
\hline
\textbf{People} on a snowy hill.                                  & The people are \textbf{women}.                      & Neutral  \\
\textbf{People} attempting to move X-ray machine down the street. & \textbf{Men} working in healthcare support service. & Neutral  \\
\textbf{People }wait for their luggage at an airport.             & \textbf{Men }are waiting for their luggage.         & Neutral  \\
\hline
\end{tabular}
\caption{\label{tab1} A typical strategy to neutralize the hypothesis in SNLI dataset. The subject word \textit{"people"} is replaced by a specific gender word such as \textit{"man"}, or \textit{"woman"} to obtain neutral examples.}
\end{table*}

\section{Related works}

\subsection{Dataset artifact detection}


Several strategies have been developed to uncover biases in NLI datasets. In 2018, a pioneering study by Poliak et al. introduced a hypothesis-only baseline, wherein premises are excluded from the dataset for both training and evaluation. Given the distinctive manner in which NLI baseline datasets are generated through crowdsourcing, a related study suggests that models can achieve satisfactory performance even when evaluated on hypothesis-only baseline datasets \cite{gururangan2018annotation}.

Another approach involves the creation of contrast examples, achieved by perturbing input in various ways and assessing the model's classification of these examples. This methodology highlights regions where the model might make incorrect decisions, providing an avenue for addressing potential dataset artifacts contributing to inaccuracies \cite{gardner2020evaluating}.

An alternative technique is the checklist test, comprising a series of assessments covering a diverse range of common behaviors that NLP models should handle. The model is then evaluated on these trials to identify potential artifacts \cite{ribeiro2020beyond}.

Further investigations are dedicated to adversarial challenge sets \cite{nie2019adversarial, mccoy2019right, wallace2019universal}. For instance, McCoy et al. introduced adversarial challenges to investigate syntactic heuristics, while Wallace et al. focused on trigger patterns causing incorrect predictions in NLI models. In addition, statistical tests are also employed to detect potential artifacts, such as the "competency problems" framework \cite{gardner2021competency}

\subsection{Dataset artifact mitigation}
Various strategies exist for addressing artifacts in datasets used for NLP models. In cases where the available training data is insufficient, one approach involves supplementing it with "adversarial datasets." These datasets are generated by perturbing and transforming the original corpus on various levels, providing additional sets for NLP model learning \cite{morris2020textattack}. Another approach that aids the model in better handling the unique challenges present in the dataset is inoculating a model by fine-tuning it on a small, problem-specific dataset \cite{liu2019inoculation}. Additionally, focusing on contradiction-word bias and word-overlapping bias is an effective strategy to address dataset artifacts, as shown by Zhou and Bansal. Specifically, the proposed method explores both data-level and model-level debiasing methods to enhance the robustness of models against lexical dataset biases \cite{zhou2020towards}.

\section{Materials and methods}
\subsection{Dataset}
In this work, we employ the widely used SNLI dataset to examine and mitigate dataset artifacts. The dataset, introduced by the Stanford NLP group, is commonly used for evaluating the performance of machine learning models in tasks related to textual entailment. Created using crowdsourcing, annotators provided labels for sentence pairs. Specifically, the premise sentences are captions from the Flickr30k corpus, and the hypotheses were generated through crowd workers following specific rules. The original SNLI dataset contains over 570,000 labeled sentence pairs distributed into three main subsets: training, development (validation), and test sets, with around 550,000, 10,000, and 10,000 examples, respectively. Each sentence pair in the dataset is labeled with one of three categories: "entailment" (the hypothesis logically follows from the premise), "neutral" (no clear logical relationship exists between the premise and hypothesis), or "contradiction" (the hypothesis contradicts the premise) \cite{bowman2015large}.

\subsection{Pre-trained language model and learning optimization}

Over the past few years, ELECTRA, or Efficiently Learning an Encoder that Classifies Token Replacements Accurately, has become increasingly popular in the field of NLP. ELECTRA was introduced by Clark et al. in 2020 as a transformer-based model specifically tailored for efficient pre-training and effective downstream task performance through fine-tuning. Pre-training ELECTRA also involves replacing a portion of the input text with incorrect or "masked" tokens, similar to other language models like BERT. However, the key difference that makes ELECTRA more efficient is that the problem has been reframed as a binary classification problem, where the model is tasked with identifying whether each token in a sequence is "real" or "replaced" \cite{clark2020electra}.

Numerous studies have shown that it significantly enhances performance on NLP tasks, including NLI and question answering. ELECTRA-small is a compact version of the original ELECTRA model that requires fewer computing resources for both training and optimization. Despite its reduced size, the ELECTRA-small model can deliver competitive performance to the standard ELECTRA model in specific tasks. Consequently, it is the preferred choice for use cases with constrained computer resources. In this work, we employ a learning optimization approach using the pre-trained ELECTRA-small from the Huggingface Transformers repository \cite{wolf2020transformers}. Specifically, we fine-turn the model with the provided API of  \verb|AutoModelForSequenceClassification| for a three-class classification problem. In this study, all experimental models are fine-tuned with approximately 12,800 steps, each step containing a batch size of 256. Model checkpoints are recorded for every 500 steps. Subsequently, the development set is used to determine the best model checkpoint to compute the final performances based on the remaining test set.

\section{Analysis of dataset artifacts}
\subsection{Hypothesis-only baseline model}

In our study, we examine dataset artifacts through the utilization of a hypothesis-only baseline model as described previously \cite{poliak2018hypothesis}. The approach involves excluding all premise sentences from the dataset and exclusively employing hypothesis sentences for both training and evaluation purposes. To implement this, we leverage the \verb|datasets.load_dataset| function from the datasets package to load the SNLI dataset. Subsequently, we partition the dataset into training, developing, and testing sets and remove premise sentences for all subsets, retaining only the hypothesis sentences. The hypothesis-only training set is then utilized to fine-tune the ELECTRA-small pre-trained model. Following the model's training, we optimize the models based on the developing set performance to select the optimal checkpoint before assessing its performance on the hypothesis-only testing set. Additionally, we conduct a comparative analysis between the performance of the hypothesis-only model and a model trained on the complete SNLI dataset, encompassing both premise and hypothesis sentences. This comparative evaluation aims to identify the impact of premise sentence removal on the model's overall performance and also to establish a baseline for evaluating our strategies for artifact mitigation later.

\begin{table}
\centering
\begin{tabular}{lrl} 
\hline
\textbf{Models}                 & \multicolumn{1}{l}{\textbf{Accuracy}} & \textbf{Dataset}  \\ 
\hline
Premise and hypothesis & 88.32                                             & SNLI              \\
Hypothesis-only        & 69.37                                             & SNLI              \\
\hline
\end{tabular}
\caption{\label{tab2} ELECTRA small performances on the SNLI dataset. "Premise and hypothesis" indicates the model trained with complete data, while "hypothesis-only" indicates the model trained solely on the hypothesis. Performances are shown in percentages.}
\end{table}

Table~\ref{tab2} provides a performance comparison between two models: one trained on both premise and hypothesis sentences and the other trained only on hypothesis sentences. The accuracy values indicate how well each model performed on the SNLI testing set. The accuracy of the standard premise and hypothesis model is high, as expected, at 88.32\%. Surprisingly, the model trained only on the hypothesis also achieves relatively good performance, with 69.37\% accuracy. This is much more impressive than the expected random guess accuracy of 1/3, indicating the existence of spurious patterns present in the datasets. As a result, the hypothesis-only model can exploit these signals to achieve high performance without learning the relationships between premise and hypothesis.

\subsection{Statistical testing}

\begin{table*}
\centering
\begin{tabular}{>{\hspace{0pt}}m{0.20\linewidth}>{\hspace{0pt}}m{0.1\linewidth}>{\RaggedLeft\hspace{0pt}}m{0.14\linewidth}>{\RaggedLeft\hspace{0pt}}m{0.102\linewidth}>{\RaggedLeft\hspace{0pt}}m{0.177\linewidth}>{\hspace{0pt}}m{0.123\linewidth}} 
\hline
\textbf{Type}                & \textbf{Word}    & \multicolumn{1}{>{\hspace{0pt}}m{0.144\linewidth}}{\textbf{Entailment}} & \multicolumn{1}{>{\hspace{0pt}}m{0.102\linewidth}}{\textbf{Neutral}} & \multicolumn{1}{>{\hspace{0pt}}m{0.177\linewidth}}{\textbf{Contradiction}} &\textbf{ P-value}   \\ 
\hline
Expected proportion &         & 33.39                                                          & 33.27                                                       & 33.35                                                             & 1.0       \\ 
\hline
Main subject (noun) & man     & 31.66                                                          & 33.21                                                       & 35.13                                                             & 6.0e-9    \\
Main subject (noun) & women   & 29.92                                                          & 31.66                                                       & 38.41                                                             & 4.0e-80   \\
Main subject (noun) & people  & \textbf{41.93}                                                          & 28.96                                                       & 29.12                                                             & 1.0e-inf  \\
Main subject (noun) & men     & 30.31                                                          & 33.16                                                       & 36.52                                                             & 4.6e-14   \\
Main subject (noun) & boy     & 29.51                                                          & 34.70                                                       & 35.79                                                             & 1.9e-14   \\ 
\hline
Main verb           & is      & 34.64                                                          & 35.94                                                       & 29.42                                                             & 7.3e-81   \\
Main verb           & are     & 39.16                                                          & 33.29                                                       & 27.55                                                             & 1.2e-132  \\
Main verb           & playing & 34.88                                                          & 30.62                                                       & 34.50                                                             & 1.2e-17   \\
Main verb           & wearing & 37.76                                                          & 29.89                                                       & 32.35                                                             & 2.3e-36   \\
Main verb           & sitting & 29.34                                                          & 16.17                                                       & \textbf{54.49}                                                             & 1.0e-inf  \\
\hline
\end{tabular}
\caption{\label{tab3} Test statistics results including label proportions corresponding to three classes and p-values for top 10 common subject nouns and main verbs. The first row is the expected proportions of three classes across the dataset. Extreme values are highlighted.}
\end{table*}

Although the hypothesis-only approach is an effective means for detecting and measuring dataset artifacts by examining model performances \cite{poliak2018hypothesis, tsuchiya2018performance, gururangan2018annotation}, the key disadvantage of this approach is the lack of interpretability. In other words, the method cannot show us why hypothesis-only models are performing so well and cannot pinpoint biases. Motivated by the observation that hypothesis-only models can predict textual entailment accurately without seeing the premises, we hypothesize that the vocabulary or word choice in the hypothesis may be associated with its labels. To address this issue, we propose a simple yet effective statistical testing procedure to unveil biases behind hypothesis-only models.

The proposed method firstly performs part-of-speech tagging parsing to extract the syntactic information of the sentences. It then applies rule-based filters to extract the most meaningful words in the sentences such as the main subjects, or main verbs. Finally, a statistical test for the association between the extracted words and hypothesis labels is carried out with $\chi^2$ goodness of fit test given the null hypothesis is the expected proportion of each label class.

To implement this idea, we apply the procedure to extract the main subject (noun) and main verb for each hypothesis example in the training set and collect the corresponding labels. Examples in which our syntactic parsing fails to extract information are excluded from the analysis. We then perform statistical testing on the top 5 common nouns and verbs against the expected proportions. The top popular nouns and verbs tested are \textit{"man," "woman," "people," "men," "boy"} and \textit{"is," "are," "playing," "wearing," "sitting"} respectively.

Results of the tests are shown in Table~\ref{tab3} and visualized in appendix Figure~\ref{fig1}, \ref{fig2}. The first subplot being the expected proportion across the dataset and the remaining 5 subplots corresponding to the top 5 common words for each type. In general, there is a clear association between label classes and tested words. For example, the word \textit{"people"} likely go with "\textit{entailment}" (41.93\%) why other subjects like \textit{"man, women, boy, men"} tends to be accompanied with \textit{"contradiction"}. Similarly, the tests for main verbs also reveal that verbs like \textit{"wearing, are"} tend to be \textit{"entailment"} while \textit{"sitting"} strongly associated with "contradiction" (54.49\%). These evidences indicate that vocabulary biases toward a specific label, i.e., knowing who is doing what, indeed carry information about the true labels of the hypothesis, and also help us to explain why the hypothesis-only model performs much better than expected by chance.

\section{Proposed approaches}

The identification of vocabulary biases naturally prompts us to the exploration of data augmentation strategies. The key idea is introducing perturbations that may aid in balancing the vocabulary distributions for the three labels. To assess this idea, we employ several types of data augmentations commonly utilized in NLP, such as character perturbations and diverse word replacement methods. Given that the biases originate from the hypotheses, our augmentation procedures are exclusively applied to this segment of the provided examples.

Given the typically large size of NLI datasets, our objective is to develop a fully automatic system for generating augmented data. To this end, we adopt Natural Language Processing Augmentation or \verb|nlpaug| package \footnote{\url{https://github.com/makcedward/nlpaug}} \cite{ma2019nlpaug} to generate augmented text data. \verb|nlpaug| is a widely used Python library designed for data augmentation in NLP. \verb|nlpaug| offers a variety of techniques, from simple rule-based approaches to more sophisticated methods like back translation and contextual word embedding. In addition, various augmentation methods implemented in the package are provided with simple and consistent API. 

In detail, this study implements five distinct NLP augmentation strategies, comprising one at the character level and four at the word level, with the aim of enhancing model accuracy and alleviating dataset biases. For character-level augmentation, the \verb|nlpaug.augmenter.char.RandomCharAug| class with \verb|action="substitute"| is applied to generate random character replacements for the given hypothesis. At a higher level, random word augmentation involves replacing words randomly using sophisticated methods that identify suitable replacements through similar embeddings, word synonym exploitation, and matched \verb|tf-idf| distribution. Specifically, the \verb|nlpaug.augmenter.word.WordEmbsAug| is employed with the \verb|word2vec| model \cite{mikolov2013efficient} as the input embedding for similar word finding. For synonym word augmentation, the \verb|nlpaug.augmenter.word.SynonymAug| is applied using two different databases: \verb|WordNet| \cite{miller1995wordnet} and \verb|The Paraphrase Database (PPDB)| \cite{ganitkevitch2013ppdb}. Regarding \verb|tf-idf| replacements, a model is first trained using the \verb|nlpaug.model.word_stats.TfIdf| API, followed by sampling using \verb|nlpaug.augmenter.word.TfIdfAug|.

Post-augmentation, the augmented data are combined with the original datasets. Subsequently, both baseline hypothesis-only models and complete models, incorporating both hypotheses and premises, are trained to evaluate the proposed approaches. Results are detailed in the following section.

\section{Experimental results}

\begin{table*}
\centering
\begin{tabular}{p{6cm}p{6cm}p{2cm}}  
\hline
\textbf{Augmentation approaches} & \multicolumn{1}{l}{\textbf{Premise and hypothesis}} & \multicolumn{1}{l}{\textbf{Hypothesis}-\textbf{only}}  \\ 
\hline
No - baseline                    & 88.32                                               & 69.37                                                  \\
Character                        & 88.74                                               & 69.29                                                  \\
Word embeding (word2vec)         & 88.62                                               & \textbf{68.23}                                         \\
Word sysnonym (PPDB)             & \textbf{88.98}                                      & 69.59                                                  \\
Word sysnonym (wordnet)          & 88.84                                               & 69.24                                                  \\
Word distribution (tf-idf)       & 88.56                                               & 68.81                                                  \\
\hline
\end{tabular}
\caption{\label{tab4} ELECTRA small performances on the SNLI dataset with various data augmentation techniques. "Premise and hypothesis" indicates the model trained with complete data, while "hypothesis-only" indicates the model trained solely on the hypothesis. Performances are shown in percentages. The top performance for each model type is highlighted.}
\end{table*}

The performance of ELECTRA small on the SNLI dataset with various data augmentation techniques is presented in Table~\ref{tab4}. In general, we anticipate an effective solution that enhances the overall model performance while diminishing the efficiency of the hypothesis-only model. As indicated in Table~\ref{tab4}, all augmentation approaches result in performance improvements for the complete models, while four out of five approaches show reductions in performance for the hypothesis-only model.

The most effective approach for mitigating biases in the SNLI dataset is word replacement by \verb|word2vec| similarity, leading to a reduction in hypothesis-only accuracy of 1.14\% from 69.37\% to 68.23\%. Following closely is \verb|tf-idf| distribution sampling, which results in a reduction of 0.56\%. In terms of complete model optimization, augmentation by word synonym from \verb|BPDP| proves to be the most successful approach, demonstrating an improvement of 0.66\% from 88.32\% to 88.98\%. Additionally, word synonym augmentation by \verb|WordNet| and character-level augmentation are also effective, achieving accuracies of 88.84\% and 88.74\%, respectively.



\section{Conclusion and future works}

In this study, we address a prevalent issue in NLP - dataset artifacts, with a specific focus on the Natural Language Inference (NLI) domain. We identify these artifacts in the SNLI dataset using a hypothesis-only baseline model by fine-tuning the pre-trained ELECTRA model. Additionally, we introduce and apply an innovative statistical testing procedure to uncover the sources of these artifacts. Our investigation highlights that biases originate from the vocabulary distribution in the hypothesis sets. To mitigate these dataset artifacts, we propose several data augmentation techniques designed to shift the distribution.

Through evaluations on both complete models and hypothesis-only models, we observe that data augmentation is indeed effective in enhancing the performance of complete models and reducing biases in their hypothesis-only counterparts.

While our findings are promising, there are opportunities for future improvements. The current implementation heavily relies on the \verb|nlpagu| library, which may not offer the highest flexibility for the NLI domain. One limitation is that the current augmentation approach does not consider the results of the proposed statistical tests. In the future, we believe that augmentations targeting highly biased words could lead to improved performance in both model accuracy and artifact mitigation.

\section*{Ethics Statement}

Not applicable.

\section*{Acknowledgements}

I would like to thank my beloved person, My Phan, for always supporting my research.

\bibliography{custom}

\begin{thebibliography}{21}
\expandafter\ifx\csname natexlab\endcsname\relax\def\natexlab#1{#1}\fi

\bibitem[{Bowman et~al.(2015)Bowman, Angeli, Potts, and Manning}]{bowman2015large}
Samuel~R Bowman, Gabor Angeli, Christopher Potts, and Christopher~D Manning. 2015.
\newblock A large annotated corpus for learning natural language inference.
\newblock \emph{arXiv preprint arXiv:1508.05326}.

\bibitem[{Clark et~al.(2020)Clark, Luong, Le, and Manning}]{clark2020electra}
Kevin Clark, Minh-Thang Luong, Quoc~V Le, and Christopher~D Manning. 2020.
\newblock Electra: Pre-training text encoders as discriminators rather than generators.
\newblock \emph{arXiv preprint arXiv:2003.10555}.

\bibitem[{Ganitkevitch et~al.(2013)Ganitkevitch, Van~Durme, and Callison-Burch}]{ganitkevitch2013ppdb}
Juri Ganitkevitch, Benjamin Van~Durme, and Chris Callison-Burch. 2013.
\newblock Ppdb: The paraphrase database.
\newblock In \emph{Proceedings of the 2013 Conference of the North American Chapter of the Association for Computational Linguistics: Human Language Technologies}, pages 758--764.

\bibitem[{Gardner et~al.(2020)Gardner, Artzi, Basmova, Berant, Bogin, Chen, Dasigi, Dua, Elazar, Gottumukkala et~al.}]{gardner2020evaluating}
Matt Gardner, Yoav Artzi, Victoria Basmova, Jonathan Berant, Ben Bogin, Sihao Chen, Pradeep Dasigi, Dheeru Dua, Yanai Elazar, Ananth Gottumukkala, et~al. 2020.
\newblock Evaluating models' local decision boundaries via contrast sets.
\newblock \emph{arXiv preprint arXiv:2004.02709}.

\bibitem[{Gardner et~al.(2021)Gardner, Merrill, Dodge, Peters, Ross, Singh, and Smith}]{gardner2021competency}
Matt Gardner, William Merrill, Jesse Dodge, Matthew~E Peters, Alexis Ross, Sameer Singh, and Noah~A Smith. 2021.
\newblock Competency problems: On finding and removing artifacts in language data.
\newblock \emph{arXiv preprint arXiv:2104.08646}.

\bibitem[{Gururangan et~al.(2018)Gururangan, Swayamdipta, Levy, Schwartz, Bowman, and Smith}]{gururangan2018annotation}
Suchin Gururangan, Swabha Swayamdipta, Omer Levy, Roy Schwartz, Samuel~R Bowman, and Noah~A Smith. 2018.
\newblock Annotation artifacts in natural language inference data.
\newblock \emph{arXiv preprint arXiv:1803.02324}.

\bibitem[{Liu et~al.(2019)Liu, Schwartz, and Smith}]{liu2019inoculation}
Nelson~F Liu, Roy Schwartz, and Noah~A Smith. 2019.
\newblock Inoculation by fine-tuning: A method for analyzing challenge datasets.
\newblock \emph{arXiv preprint arXiv:1904.02668}.

\bibitem[{Ma(2019)}]{ma2019nlpaug}
Edward Ma. 2019.
\newblock Nlp augmentation.
\newblock https://github.com/makcedward/nlpaug.

\bibitem[{MacCartney and Manning(2008)}]{maccartney2008modeling}
Bill MacCartney and Christopher~D Manning. 2008.
\newblock Modeling semantic containment and exclusion in natural language inference.
\newblock In \emph{Proceedings of the 22nd International Conference on Computational Linguistics (Coling 2008)}, pages 521--528.

\bibitem[{McCoy et~al.(2019)McCoy, Pavlick, and Linzen}]{mccoy2019right}
R~Thomas McCoy, Ellie Pavlick, and Tal Linzen. 2019.
\newblock Right for the wrong reasons: Diagnosing syntactic heuristics in natural language inference.
\newblock \emph{arXiv preprint arXiv:1902.01007}.

\bibitem[{Mikolov et~al.(2013)Mikolov, Chen, Corrado, and Dean}]{mikolov2013efficient}
Tomas Mikolov, Kai Chen, Greg Corrado, and Jeffrey Dean. 2013.
\newblock Efficient estimation of word representations in vector space.
\newblock \emph{arXiv preprint arXiv:1301.3781}.

\bibitem[{Miller(1995)}]{miller1995wordnet}
George~A Miller. 1995.
\newblock Wordnet: a lexical database for english.
\newblock \emph{Communications of the ACM}, 38(11):39--41.

\bibitem[{Morris et~al.(2020)Morris, Lifland, Yoo, Grigsby, Jin, and Qi}]{morris2020textattack}
John~X Morris, Eli Lifland, Jin~Yong Yoo, Jake Grigsby, Di~Jin, and Yanjun Qi. 2020.
\newblock Textattack: A framework for adversarial attacks, data augmentation, and adversarial training in nlp.
\newblock \emph{arXiv preprint arXiv:2005.05909}.

\bibitem[{Nie et~al.(2019)Nie, Williams, Dinan, Bansal, Weston, and Kiela}]{nie2019adversarial}
Yixin Nie, Adina Williams, Emily Dinan, Mohit Bansal, Jason Weston, and Douwe Kiela. 2019.
\newblock Adversarial nli: A new benchmark for natural language understanding.
\newblock \emph{arXiv preprint arXiv:1910.14599}.

\bibitem[{Poliak et~al.(2018)Poliak, Naradowsky, Haldar, Rudinger, and Van~Durme}]{poliak2018hypothesis}
Adam Poliak, Jason Naradowsky, Aparajita Haldar, Rachel Rudinger, and Benjamin Van~Durme. 2018.
\newblock Hypothesis only baselines in natural language inference.
\newblock \emph{arXiv preprint arXiv:1805.01042}.

\bibitem[{Ribeiro et~al.(2020)Ribeiro, Wu, Guestrin, and Singh}]{ribeiro2020beyond}
Marco~Tulio Ribeiro, Tongshuang Wu, Carlos Guestrin, and Sameer Singh. 2020.
\newblock Beyond accuracy: Behavioral testing of nlp models with checklist.
\newblock \emph{arXiv preprint arXiv:2005.04118}.

\bibitem[{Tsuchiya(2018)}]{tsuchiya2018performance}
Masatoshi Tsuchiya. 2018.
\newblock Performance impact caused by hidden bias of training data for recognizing textual entailment.
\newblock \emph{arXiv preprint arXiv:1804.08117}.

\bibitem[{Wallace et~al.(2019)Wallace, Feng, Kandpal, Gardner, and Singh}]{wallace2019universal}
Eric Wallace, Shi Feng, Nikhil Kandpal, Matt Gardner, and Sameer Singh. 2019.
\newblock Universal adversarial triggers for attacking and analyzing nlp.
\newblock \emph{arXiv preprint arXiv:1908.07125}.

\bibitem[{Williams et~al.(2017)Williams, Nangia, and Bowman}]{williams2017broad}
Adina Williams, Nikita Nangia, and Samuel~R Bowman. 2017.
\newblock A broad-coverage challenge corpus for sentence understanding through inference.
\newblock \emph{arXiv preprint arXiv:1704.05426}.

\bibitem[{Wolf et~al.(2020)Wolf, Debut, Sanh, Chaumond, Delangue, Moi, Cistac, Rault, Louf, Funtowicz et~al.}]{wolf2020transformers}
Thomas Wolf, Lysandre Debut, Victor Sanh, Julien Chaumond, Clement Delangue, Anthony Moi, Pierric Cistac, Tim Rault, R{\'e}mi Louf, Morgan Funtowicz, et~al. 2020.
\newblock Transformers: State-of-the-art natural language processing.
\newblock In \emph{Proceedings of the 2020 conference on empirical methods in natural language processing: system demonstrations}, pages 38--45.

\bibitem[{Zhou and Bansal(2020)}]{zhou2020towards}
Xiang Zhou and Mohit Bansal. 2020.
\newblock Towards robustifying nli models against lexical dataset biases.
\newblock \emph{arXiv preprint arXiv:2005.04732}.

\end{thebibliography}
\bibliographystyle{acl_natbib}

\appendix

\section{Appendix figures}
\label{sec:appendix}

\begin{figure*}[htb]
    \centering
    \includegraphics[width=0.9\textwidth]{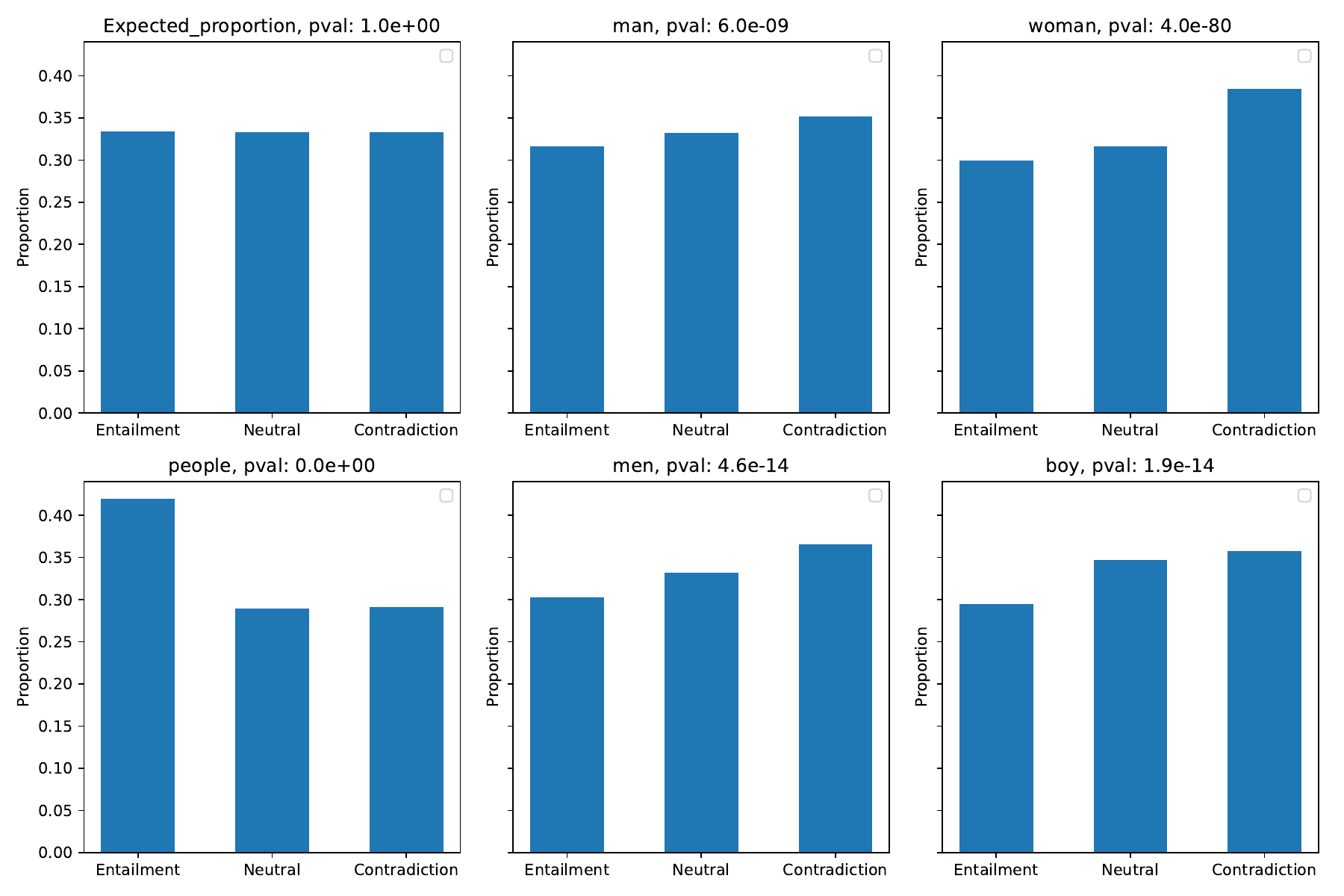}
    \caption{Statistics results including label proportions corresponding to three classes and p-values for top 5 common subject nouns. The first row is the expected proportions of three classes across the dataset.}
    \label{fig1}
\end{figure*}

\begin{figure*}[htb]
    \centering
    \includegraphics[width=0.9\textwidth]{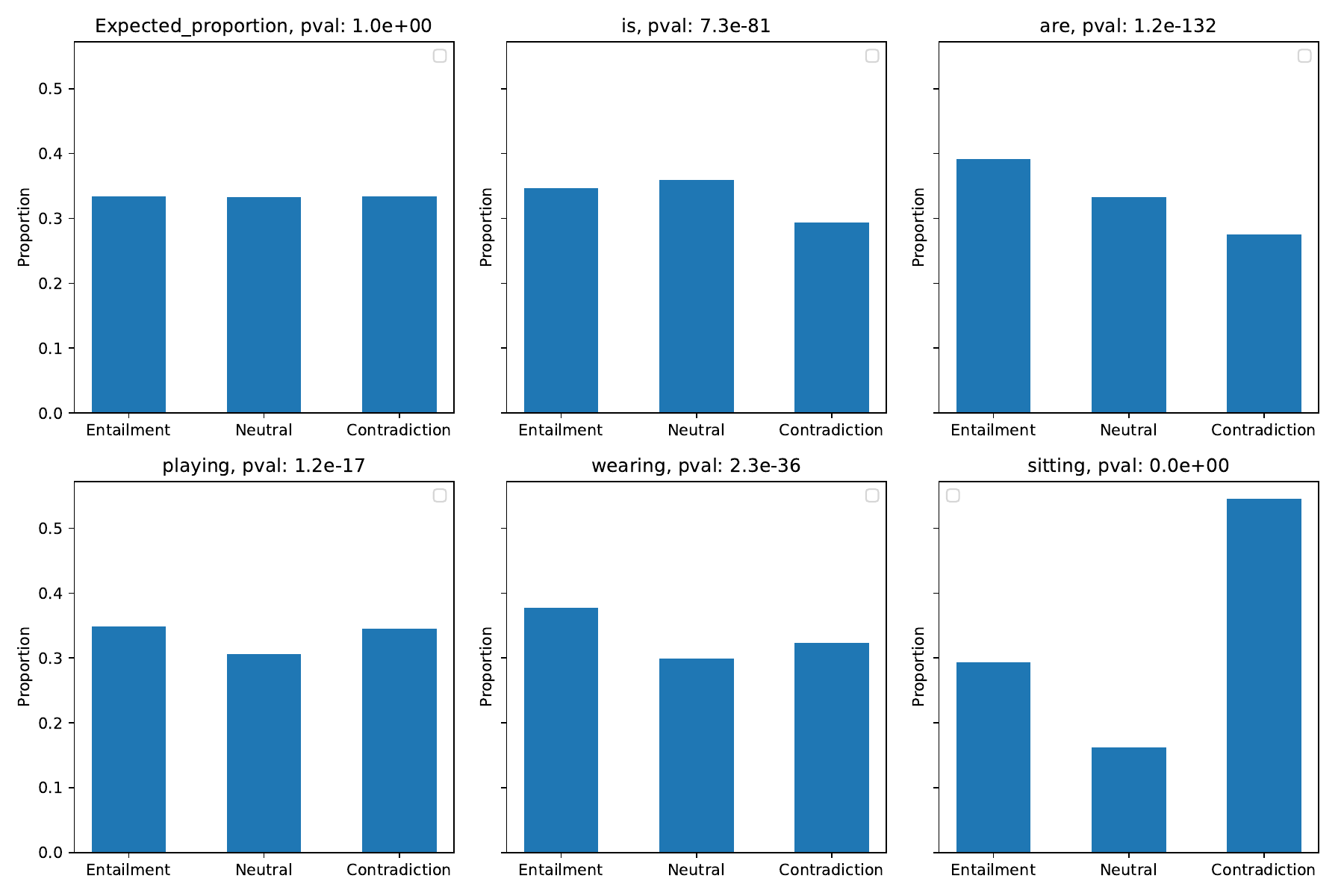}
    \caption{Statistics results including label proportions corresponding to three classes and p-values for top 5 common main verbs. The first row is the expected proportions of three classes across the dataset.}
    \label{fig2}
\end{figure*}

\end{document}